\documentclass[11pt,a4paper]{article}
\usepackage[hyperref]{emnlp2020}
\usepackage{times}
\usepackage{latexsym}

\usepackage{graphicx}
\usepackage{subfigure}
\usepackage{microtype}
\usepackage{amsmath}
\usepackage{multirow}
\usepackage{booktabs}
\usepackage{graphviz}
\usepackage{amssymb}
\usepackage{array}
\usepackage{enumitem}
\newcolumntype{P}[1]{>{\centering\arraybackslash}p{#1}}
\newcolumntype{M}[1]{>{\centering\arraybackslash}m{#1}}

\usepackage{amsmath,amsfonts,bm}

\def\Figref#1{Figure~\ref{#1}}

\def\secref#1{section~\ref{#1}}

\def\twosecrefs#1#2{sections \ref{#1} and \ref{#2}}

\def\eqref#1{equation~\ref{#1}}
\def\Eqref#1{Equation~\ref{#1}}

\def\floor#1{\lfloor #1 \rfloor}
\def\1{\bm{1}}

\DeclareMathAlphabet{\mathsfit}{\encodingdefault}{\sfdefault}{m}{sl}
\SetMathAlphabet{\mathsfit}{bold}{\encodingdefault}{\sfdefault}{bx}{n}

\newcommand{\R}{\mathbb{R}}

\newcommand{\softmax}{\mathrm{softmax}}

\usepackage{url}
\usepackage{bm}

\newcommand{\hide}[1]{} 
\newcommand{\vpara}[1]{\vspace{0.1in}\noindent\textbf{#1 }}

\newcommand{\beq}[1]{\small \begin{equation}#1\end{equation}\normalsize}

\newcommand{\besp}[1]{\begin{split}#1\end{split}}

\def\method{Block\textsc{Bert}}
\def\shortmethod{Block\textsc{Bert}}
\def\sparse{Sparse\textsc{Bert}}

\aclfinalcopy 

\title{Blockwise Self-Attention for Long Document Understanding}

\author{
    Jiezhong Qiu$^{1}$\Thanks{This work was partially done when the first author was an intern at Facebook AI. Code is available at \url{https://github.com/xptree/BlockBERT}}
    , Hao Ma$^{2}$, Omer Levy$^{2}$, Wen-tau Yih$^{2}$, Sinong Wang$^{2}$, Jie Tang$^{1}$\\
    $^{1}$Department of Computer Science and Technology, Tsinghua University \\
    $^{2}$Facebook AI \\
    \texttt{qiujz16@mails.tsinghua.edu.cn} \\
    \texttt{\{haom,omerlevy,scottyih,sinongwang\}@fb.com}\\
    \texttt{jietang@tsinghua.edu.cn}

}
\date{}

\begin{document}
\maketitle

\begin{abstract}
We present \method{}, a lightweight and efficient BERT model 
for better modeling long-distance dependencies.
Our model extends BERT by introducing sparse block structures 
into the attention matrix to reduce both memory consumption and training/inference time, 
which also enables attention heads to capture either short- or long-range contextual information.
We conduct experiments on language model pre-training and 
several benchmark question answering datasets with various paragraph lengths. 
\method{} 
uses 18.7-36.1\% less memory and 12.0-25.1\% less time to learn the model. 
During testing, \method{} saves 27.8\% inference time, while having
comparable and sometimes better prediction accuracy, compared to an advanced BERT-based model, RoBERTa.
\end{abstract}

\section{Introduction}

Recent emergence of the \emph{pre-training} and \emph{fine-tuning} paradigm, exemplified by methods like ELMo~\citep{peters2018deep}, GPT-2/3~\citep{radford2019language, brown2020language}, BERT~\citep{devlin2019bert}, XLNet~\citep{yang2019xlnet}, RoBERTa~\citep{liu2019roberta} and ALBERT~\cite{lan2019albert}, has drastically reshaped the landscape of the natural language processing research.
These methods first pre-train a deep model with language model objectives using a large corpus and then fine-tune the model using in-domain
supervised data for target applications.  Despite its conceptual simplicity, this paradigm has re-established the new state-of-the-art baselines across various tasks, such as question answering~\citep{devlin2019bert}, coreference resolution~\citep{joshi2019bert}, relation extraction~\citep{soares2019matching} and text retrieval~\citep{lee2019latent, nogueira2019passage}, to name a few. 


Building such models in practice, however, is an extremely resource-intensive process.  For instance, the training of BERT-family models is notoriously expensive.  \citet{devlin2019bert} report that it takes four days to pre-train BERT-Base/BERT-Large on 4/16 Cloud TPUs.
In order to reduce the pre-training time of RoBERTa to 1 day, \citet{liu2019roberta} use 1,024 V100 GPUs.
One crucial factor contributing to the long training time is the memory consumption of these deep models, as it directly affects the batch size.
Although the fine-tuning stage is relatively inexpensive, the memory  issue still restricts the scenarios in which BERT can be used.
For instance, ``it is currently not possible to re-produce most of the BERT-Large results on the paper using a GPU with 12GB-16GB of RAM, because the maximum batch size that can fit in memory is too small.\footnote{\url{github.com/google-research/bert}}"


Although one may think that model size is the main contributor to the large memory consumption, our analysis (Section~\ref{subsec:profile})
shows that one of the main bottlenecks is actually dot-product self-attention, operated in multiple layers of Transformers~\citep{vaswani2017attention}, the building block of BERT.  As the attention operation is quadratic to the sequence length, this fundamentally limits the maximum length of the input sequence, and thus restricts the model capacity in terms of capturing long-distance dependencies.
As a result, downstream tasks have to either truncate their sequences to leading tokens~\citep{nogueira2019passage} or split their sequences with a sliding window~\citep{joshi2019spanbert, joshi2019bert}.  Ad-hoc handling of long sequences is also required in the pre-training stage, such as updating the model using only short sequences in the early stage~\citep{devlin2019bert}.


Common strategies for reducing memory consumption, unfortunately, do not work.  For instance,
shrinking the model by lowering the number of layers~$L$, attention heads~$A$, or hidden units~$H$ leads
to significant performance degradation~\citep{vaswani2017attention,devlin2019bert} and does not address
the long sequence issue.  
Alternatively, general low-memory training techniques, such as microbatching~\citep{huang2018gpipe} and 
gradient checkpointing~\citep{chen2016training} essentially trade off training time for memory consumption, 
prolongs the already lengthy training process.


In this work, we explore a different strategy, \emph{sparsifying the attention layers}, intending to design
a lightweight and effective BERT that can model long sequences in a memory-efficient way.
Our \method{} extends BERT by introducing sparse block substructures into attention matrices to reduce both 
memory consumption and the number of floating-point operations~(FLOPs), which also enables attention heads to capture either short- or long-range contextual information.
Compared to the previous method that also enforces sparsity~\cite{child2019generating}, our approach
is much simpler mathematically and very easy to implement.  More importantly, the results of experiments conducted
on several benchmark question answering datasets with various paragraph lengths show that \method{} performs
comparably or even better than the original BERT-family models, while enjoying an 18.7-36.1\% reduction in memory
usage, a 12.0-25.1\% reduction in training time, and a 27.8\% reduction in inference time. 


The rest of the paper is organized as follows.  Section~\ref{sec:background} gives a brief introduction of the BERT model, along with an in-depth analysis of its memory usage during training time.  We describe our proposed model in Section~\ref{sec:model} and contrast it with existing methods that aim for creating a lighter model.  Section~\ref{sec:exp} presents the experimental results and  ablation studies, followed by a survey of other related work in Section~\ref{sec:related} and the conclusion in Section~\ref{sec:conclusion}. 


\section{Background: Memory Bottleneck in Training BERT}
\label{sec:background}


We briefly review BERT and introduce its memory profiling in this section.  
Following the paradigm of language model pre-training 
and down-stream task fine-tuning, BERT~\citep{devlin2019bert} consists of multiple layers of bidirectional Transformers~\citep{vaswani2017attention},
where each Transformer encoder has a multi-head self-attention layer and a position-wise feed-forward layer.
Using the same notation as in~\citep{devlin2019bert}, 
we denote the number of Transformer layers by $L$, the number of hidden units by $H$, the number of attention heads by $A$, the sequence length by $N$, and the batch size by $B$. We also assume the feed-forward hidden unit size to be $4H$.\footnote{The default parameter settings for BERT-Base and BERT-Large can be found in Appendix~\ref{subsec:config}}


\subsection{Memory Profiling}
\label{subsec:profile}


Training BERT is a memory-intensive process. In order to identify the bottleneck, we follow the memory model proposed by \citet{sohoni2019low},
where memory usage throughout neural network training is categorized into three main types:
(1)~\textbf{Model memory} is used to store model parameters; (2)~\textbf{Optimizer memory} is the additional memory used by the specific learning algorithm during the process; (3)~\textbf{Activation memory} consists of the outputs of each layer, which are cached for reuse in backpropagation to compute gradients. 

Take BERT-Base training as an example.  The model has 110 million parameters, so model memory occupies 0.2~GB if parameters are stored in half-precision floating-point format~(FP16).  For Adam~\citep{kingma2014adam}, the optimizer needs additional memory to store the gradients, first moments, and second moments of model parameters.  If stored using the same precision, the optimizer memory should be three times of model memory.\footnote{In the current PyTorch Adam implementation, the first and second moments are stored in single precision. Consequently, BERT's optimizer memory~(1~GB) is five times of model memory~(0.2~GB).}
To calculate the exact size of activation memory is not trivial because it depends heavily on the implementation of the toolkit.  Instead, we measure it empirically
by training BERT-Base using Adam with a memory profiler~(more details are provided in Appendix~\ref{subsec:profiler}).

\begin{figure}[t]
     \centering
     \subfigure[BERT-Base Training Memory Profiling]{
        \includegraphics[width=.51\columnwidth]{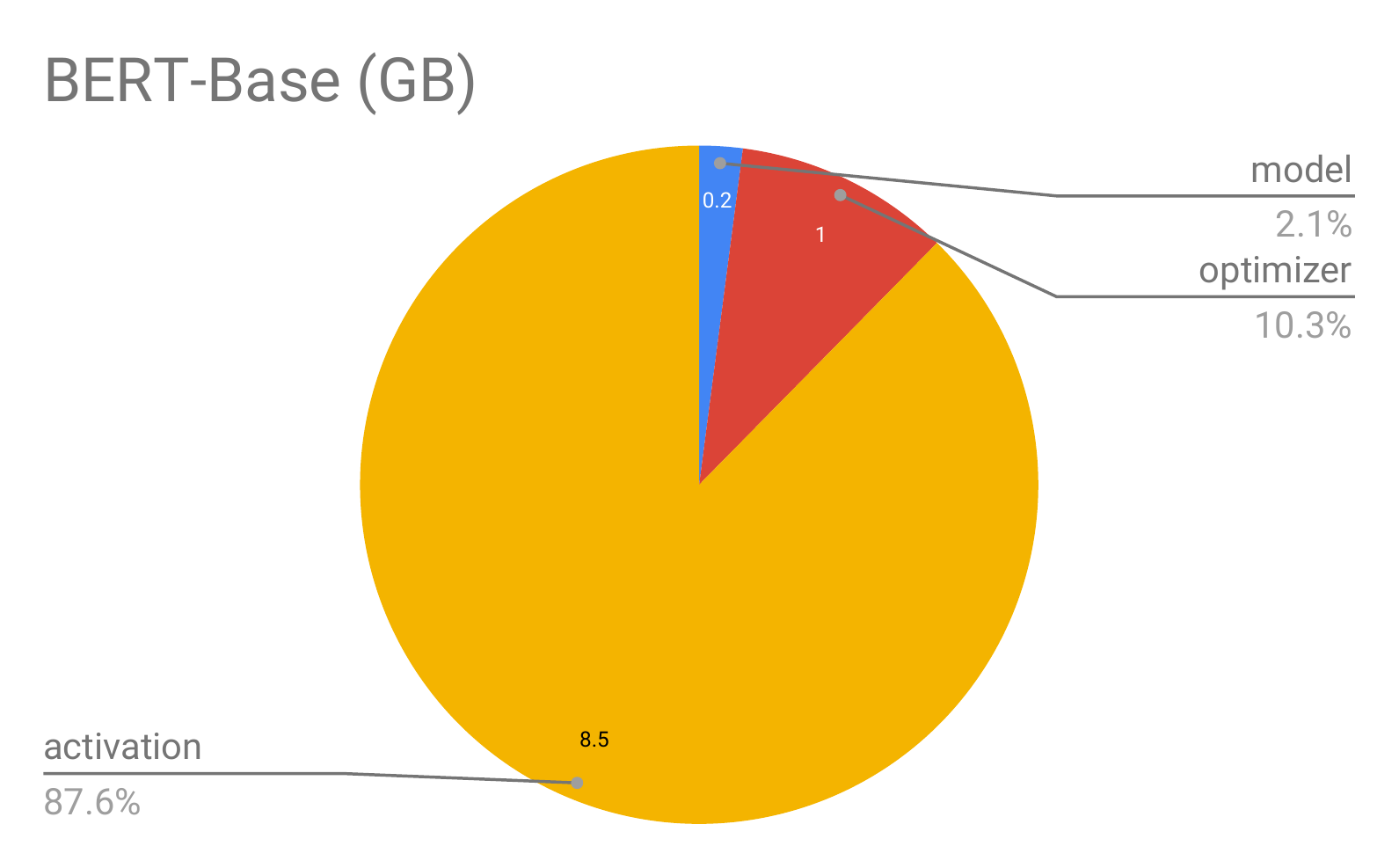}
        \label{fig:profile}
        }
     \subfigure[Regression Analysis on Activation Memory]{
        \includegraphics[width=.4\columnwidth]{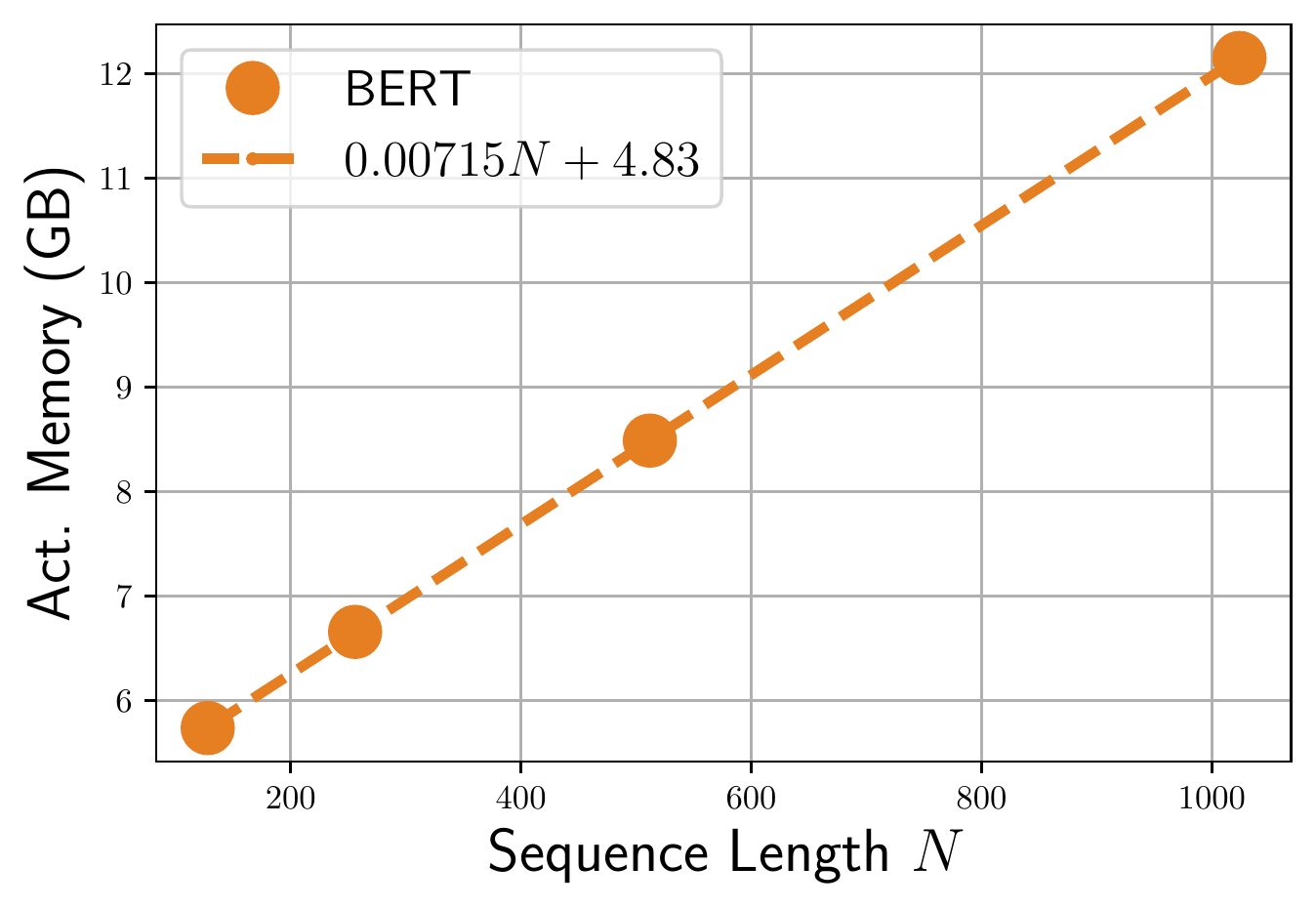}
        \label{fig:bert_regression}
        }
     \caption{Memory Profiling for BERT.}
     \label{fig:bert_memeory}
\end{figure}

We use 32 NVIDIA V100 GPUs for training.  Every single GPU thus consumes a mini-batch of size $b=B/32=8$.  
\Figref{fig:profile} shows the profiling result for a single GPU, where the model/optimizer/activation memory consumes 0.21/1.03/8.49~GB, respectively.  
We can see that activation memory accounts for the vast majority of the total GPU memory~(87.6\%) and is thus the bottleneck.
Notice that although our analysis is done on BERT-Base, it can also be generalized to BERT-Large and other  models such as RoBERTa~\citep{liu2019roberta} and  XLNet~\citep{yang2019xlnet}. 


\subsection{A Regression Analysis on Activation Memory}
\label{subsec:regression}

For BERT, or more specifically, Transformer, the activation memory corresponds to intermediate results of different layers.
It grows linearly in all the model hyper-parameters, except the sequence length $N$,
due to the attention layers.  
To quantify  the linear and quadratic components in the activation memory more clearly, we conduct a regression analysis as follows.
Assume that the activation memory~(in each GPU) is a polynomial $a_2bN^2 + a_1bN + a_0$, where $b$ is the batch size in each GPU and $a_i$~($i=0, 1, 2$) are coefficients to be determined.  If we fix the total number of tokens in a GPU to be  constant~(in our case, we fix $b\times N=4096$), we should have a linear function w.r.t. $N$, i.e., $4096a_2N + 4096a_1 + a_0$. 
We enumerate $N$ from $\{128, 256, 512, 1024\}$ in our experiments, and plot the corresponding profiled activation memory in \Figref{fig:bert_regression}.
Using ordinary least squares~(OLS), with $b\times N=4096$, the estimated linear function for activation memory is $0.00715\times N + 4.83$, where the 
first term corresponds to the $O(N^2)$ component.
When $N=512$~(i.e., $b=8$), we can see that for BERT-Base, the $O(N^2)$ component accounts for 3.66~GB, and the $O(N)$ component accounts for 4.83~GB.  When the sequence length $N$ increases to 1024~(i.e., $b=4$), the $O(N^2)$ component increases to 7.32~GB, while the $O(N)$ part is unchanged.


\subsection{Techniques for Reducing Traing Memory}
\label{subsec:tech-memory}

Observing that  activation memory is the training bottleneck, 
we discuss common memory reduction techniques below.

\vpara{Low Precision}~\citep{micikevicius2017mixed} Low precision is to use half-precision/mixed-precision for training neural networks. This technique has been widely used in Transformer training~\citep{ott2019fairseq, liu2019roberta}.  In this work, we already assume to use mixed-precision training by default, as indicated in the aforementioned analysis.

\vpara{Microbatching}~\citep{huang2018gpipe} Microbatching is to split a batch into small micro-batches~(which can be fit into memory), and then run forward and backward passes on them separately with gradients for each micro-batch accumulated. Because it runs forward/backward pass multiple times for a single batch, it trades off time for memory.

\vpara{Gradient Checkpointing}~\citep{chen2016training} Gradient checkpointing saves memory by only caching activations of a subset of layers. The un-cached activations will be recomputed during backpropagation from the latest checkpoint.  This strategy trades off time for memory by repeating computations 
and will obviously extend training time. 

\vpara{Knowledge Distillation}~\citep{hinton2015distilling} Knowledge distillation aims to compress and transfer knowledge from a teacher model to a simpler student model.  However, knowledge distillation relies on a teacher model~(which is still expensive in training time) and usually suffers from a certain degree of performance degradation.

\citep{Ding2020neurips} presents an alternative idea based on  cognitive theory to  construct a working-memory by identifying key sentences, which enables multi-step reasoning. However, common techniques are still limited in reducing both the training time and memory usage. In this paper, we investigate how to optimize the dot-product attention layers and introduce our approach next.

\section{Model: \method{}}
\label{sec:model}


Following \citep{vaswani2017attention}, the dot-product attention in Transformer is defined as:
\beq{\nonumber
  \text{Attention}(\bm{Q}, \bm{K}, \bm{V})=\softmax{\left(\frac{\bm{Q}\bm{K}^\top}{\sqrt{d}}\right)}\bm{V},
}where $\bm{Q}, \bm{K}, \bm{V} \in \R^{N\times d}$ with $N$ to be the sequence length and $d$ to be a hidden dimension. As we can see, the inner product between $\bm{Q}$ and $\bm{K}$ consumes $O(N^2)$ memory.
One simple way to reduce the memory consumption of attention is to sparsify the attention matrix.
Suppose we have a masking matrix $\bm{M}\in \{0,1\}^{N\times N}$, we define a masked version of attention as follows:

\beq{\label{eq:mask_attention}
  \text{Attention}(\bm{Q}, \bm{K}, \bm{V}, \bm{M})=\softmax{\left(\frac{\bm{Q}\bm{K}^\top}{\sqrt{d}} \odot \bm{M}\right)}\bm{V},
}with operator $\odot$ defined by
\beq{\nonumber
  (\bm{A}\odot \bm{M})_{ij} = \begin{cases} \bm{A}_{ij} & \text{if } \bm{M}_{ij} = 1 \\ -\infty & \text{if } \bm{M}_{ij}=0 \end{cases}.
}
In this work, we design $\bm{M}$ to be a \emph{sparse block matrix}, which not only reduces memory and the number of floating-point operations~(FLOPs) but also benefits
from efficient dense matrix support from deep learning frameworks, such as PyTorch and Tensorflow.
More formally, we split the length-$N$ input sequence into $n$ blocks, with each block of length $\frac{N}{n}$.\footnote{We assume $N$ can be divided by $n$. If not, we pad the input sequence to make $N$ divisible.}
The $N\times N$ attention matrix is then partitioned into $n \times n$ blocks, where each block matrix is of the size $\frac{N}{n} \times \frac{N}{n}$.
We define a sparse block matrix $\bm{M}$ by a permutation $\pi$ of $\{1, 2, \cdots, n\}$:

\beq{
  \label{eq:mask}
  \bm{M}_{ij} = \begin{cases} 1 & \text{if } \pi\left( \floor{\frac{(i-1)n}{N}+1} \right)=\floor{\frac{(j-1)n}{N}+1}, \\
    0 & \text{otherwise.}
  \end{cases}
}By writing $\bm{Q}, \bm{K}, \bm{V}$ as block matrices, such that
$\tiny
  \bm{Q}=
  \begin{bmatrix}
    \bm{Q}_1^\top &
    \cdots        &
    \bm{Q}_n^\top
  \end{bmatrix}^\top,
  \bm{K}=
  \begin{bmatrix}
    \bm{K}_1^\top &
    \cdots        &
    \bm{K}_n^\top
  \end{bmatrix}^\top\normalsize$ and
$\tiny\bm{V}=
  \begin{bmatrix}
    \bm{V}_1^\top &
    \cdots        &
    \bm{V}_n^\top
  \end{bmatrix}^\top
  \normalsize$ and pluging them into \Eqref{eq:mask_attention}, we can formally define Blockwise Attention as follows:
  
\beq{
  \label{eq:struc_attention}
  \besp{
    \text{Blockwise-Attention}&(\bm{Q}, \bm{K}, \bm{V}, \bm{M})
    \\=&\begin{bmatrix}
      \softmax{\left(\frac{\bm{Q}_1\bm{K}_{\pi(1)}^\top}{\sqrt{d}}\right)}\bm{V}_{\pi(1)} \\
      \vdots                                                                              \\
      \softmax{\left(\frac{\bm{Q}_{n}\bm{K}_{\pi(n)}^\top}{\sqrt{d}}\right)}\bm{V}_{\pi(n)}
    \end{bmatrix}.
  }}\Eqref{eq:struc_attention} only needs to compute and store $\bm{Q}_i\bm{K}_{\pi(i)}^\top$~($i=1,\cdots n$), each has size $\frac{N}{n} \times \frac{N}{n}$.  In other words, \method{}  reduces both $O(N^2)$ memory consumption and FLOPs by a factor of $n$, since $\frac{N}{n} \times \frac{N}{n} \times n = \frac{N\times N}{n}$.

\begin{figure}[htbp]
  \centering
  \includegraphics[width=.45\textwidth]{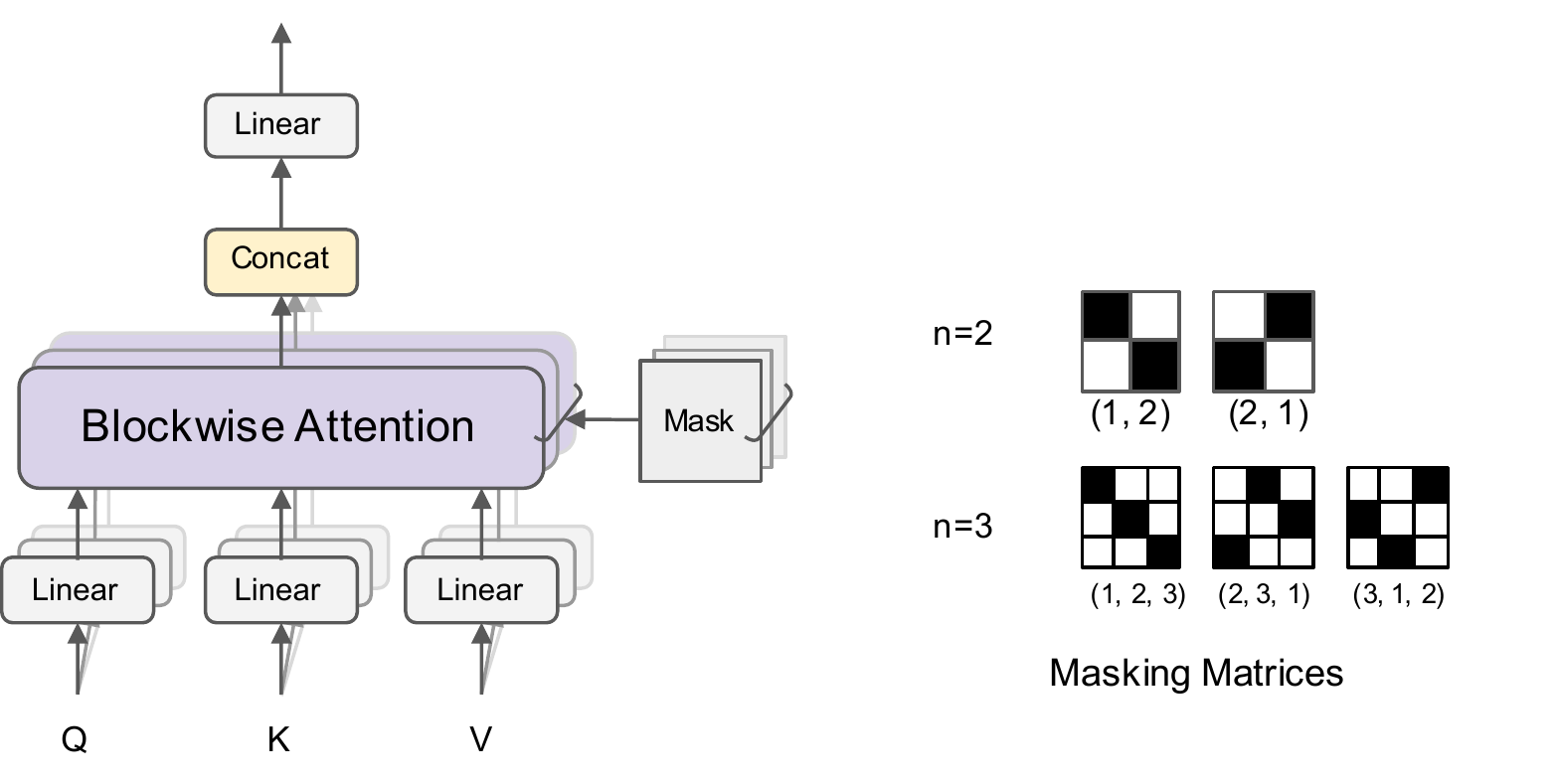}
  \caption{Architecture of Blockwise Multi-head Attention, which acts as building blocks of \method{}. The key idea is to introduce a sparse block masking matrix to the $N\times N$ attention matrix. The right panel shows the masking matrices we use when $n=2, 3$. For $n=2$, the masking matrices are defined by permutation $(1, 2)$, $(2, 1)$ and have 50\% non-zeros. For $n=3$, the masking matrices are defined by permutation $(1, 2, 3)$, $(2, 3, 1)$, and $(3, 1, 2)$ and have 33.33\% non-zeros.
  }
  \label{fig:method}
\end{figure}


\subsection{Blockwise Multi-Head Attention}

Analogous to Multi-head Attention~\citep{vaswani2017attention}, we allow queries, keys, and values to be projected multiple times and perform blockwise attentions in parallel.   Moreover, different blockwise attention heads can use different masking matrices.
The outputs of multiple heads are then concatenated and aggregated with another linear projection.
Let $A$ be the number of attention heads and $H$ the number of hidden units. \emph{Blockwise multi-head attention} is formally defined as follows:

\beq{\nonumber
  \besp{
    \text{Blockwise-Multi-head-Attention}&(\bm{Q}, \bm{K}, \bm{V}) \\
    =& \text{Concat}(\text{head}_1, \cdots \text{head}_A)\bm{W}^O,
  }}where for each head $i$, $i=1, 2, \cdots, A$, 
  \beq{\nonumber
  \text{head}_i=  \text{Blockwise-Attention}(\bm{Q}\bm{W}_i^Q, \bm{K}\bm{W}_i^K, \bm{V}\bm{W}_i^V, \bm{M}_i),
}with $d=\frac{H}{A}, \bm{W}_i^Q, \bm{W}_i^K, \bm{W}_i^V \in \mathbb{R}^{H \times d}$ and the projection matrix $\bm{W}^O \in \mathbb{R}^{H\times H}$. Each masking matrix $\bm{M}_i$ is determined by a permutation $\pi_i$ according to \Eqref{eq:mask}.
In particular, we choose $\pi$ from permutations generated by \emph{shifting one position}: $\sigma=(2, 3, \cdots, n, 1)$,
i.e., we select $\pi\in \{\sigma, \sigma^2, \cdots, \sigma^{n}\}$. For example, with 12 attention heads~($A=12$) and 2 blocks~($n=2$),
we can assign 10 heads to permutation $(1, 2)$ and the other 2 heads to permutation $(2, 1)$.
\Figref{fig:method} illustrates the blockwise multi-head attention with block number $n \in \{2,3\}$.
Blockwise sparsity captures both local and long-distance dependencies in a memory-efficiency way,
which is crucial for long-document understanding tasks.  For instance, the identity permutation, i.e., $(1, 2, \cdots, n)$,
enables each token to attend to its nearby tokens in self-attention, while other permutations allow  tokens within the same block attending to 
tokens in another block. Our proposed \method{} essentially replaces the multi-head attention layers in Transformer/BERT with blockwise multi-head attention.


\subsection{Analysis of Memory Usage Reduction}

\begin{figure}[htbp]
  \centering
  \includegraphics[width=.3\textwidth]{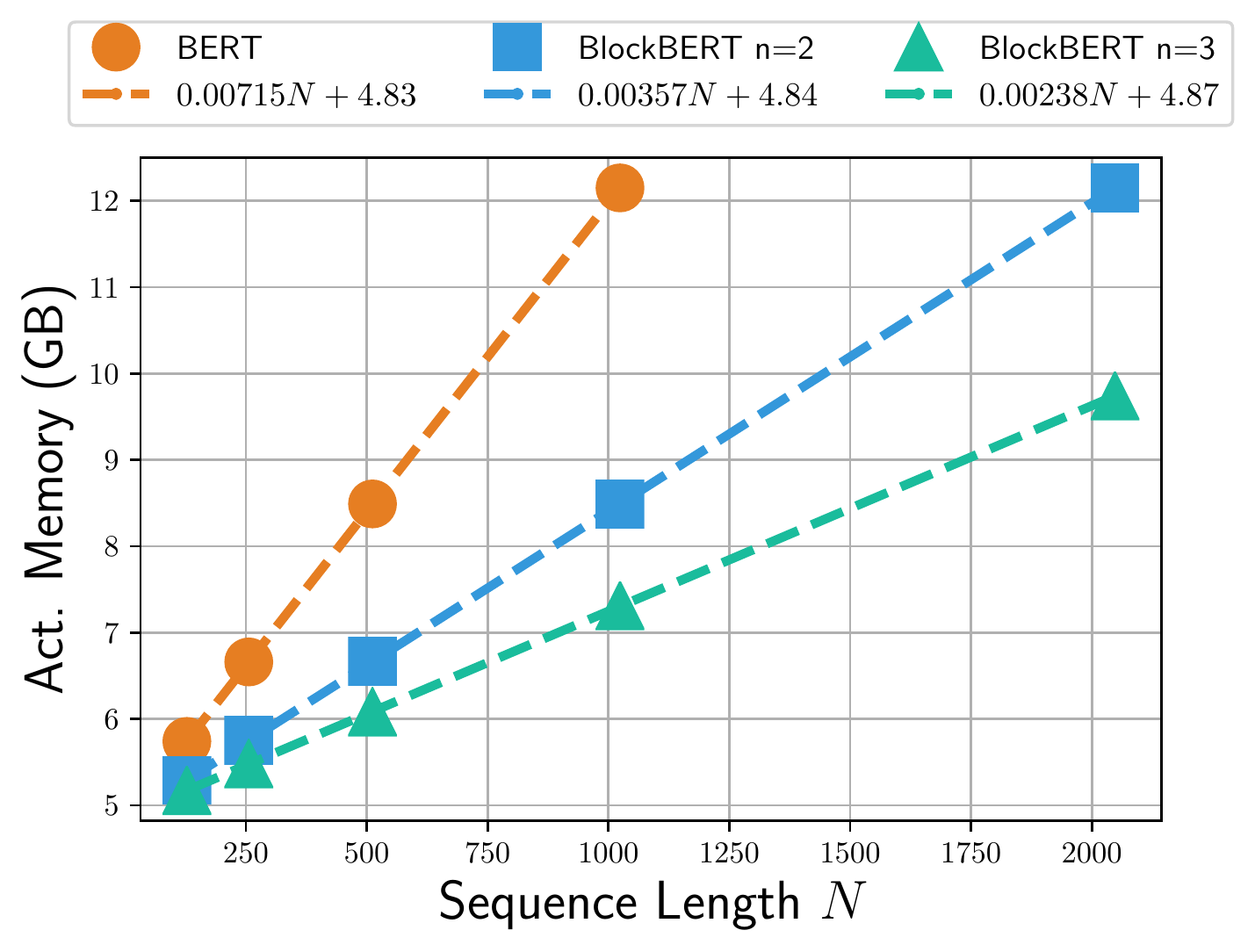}
  \caption{Regression analysis on activation memory for BERT and \method{}.}
  \label{fig:regression}
\end{figure}

To validate our claim that \method{} with $n \times n$ blocks can reduce the $O(N^2)$ memory usage by a factor of $n$,
we perform the same memory profiling as described in \twosecrefs{subsec:profile}{subsec:regression}.
Again, We fix the number of tokens in each GPU~($b \times N=4096$) and choose $N$ from $\{128, 256, 512, 1024, 2048\}$.\footnote{We use GPUs of 16~GB memory for profiling.  BERT with $N=2048$ fails due to an out-of-memory error.}
As we can see from \Figref{fig:regression} and Table~\ref{tbl:memory}, the empirical results align well with the theoretical values.
When we set the number of blocks to be 2 and 3 for \method{}, the estimated $O(N^2)$  activation memory decreases to 1/2 and 1/3 of BERT's $O(N^2)$ activation memory, respectively.
As shown in Table~\ref{tbl:pretrain}, for the sequence length $N=512$, \method{} with 2 and 3 blocks saves 18.7\%  and  23.8\% overall memory, respectively.
The saving is more significant for longer sequences.  When $N=1024$, the overall memory reduction of \method{} with 2 and 3 blocks is 27.3\% and 36.1\%, respectively.

\begin{table}[htbp]
  \small
  \centering
  \begin{tabular}{l |l| l|  l  l}
    \toprule
                          &                    &                    & \multicolumn{2}{c}{Act. Mem.~(GB)}            \\
    $N$                   & $b$                & Model              & $O(N)$                             & $O(N^2)$ \\\midrule
    \multirow{3}{*}{512}  & \multirow{3}{*}{8} & BERT               & 4.83                               & 3.66     \\
                          &                    & \shortmethod{}~n=2 & 4.84                               & 1.83     \\
                          &                    & \shortmethod{}~n=3 & 4.87                               & 1.22     \\\midrule
    \multirow{3}{*}{1024} & \multirow{3}{*}{4} & BERT               & 4.83                               & 7.32     \\
                          &                    & \shortmethod{}~n=2 & 4.84                               & 3.66     \\
                          &                    & \shortmethod{}~n=3 & 4.87                               & 2.44     \\\bottomrule
  \end{tabular}
  \normalsize
    \caption{Estimated $O(N^2)$ and $O(N)$ activation memory for BERT and \method{}. 
  } 
  \label{tbl:memory}
\end{table}

\section{Experiments}
\label{sec:exp}
We evaluate the pre-training and fine-tuning performance of \method{}.
In particular, when $n=2$, we denote 10:2 to be the configuration which assigns 10 heads to permutation $(1, 2)$ and  2 to permutation $(2, 1)$; when $n=3$, we denote 8:2:2 to be the configuration which assigns 8, 2, 2 heads to permutation $(1, 2, 3)$, $(2, 3, 1)$, and $(3, 1, 2)$, respectively.
We compare \method{} with the following baselines:

\vpara{Google BERT} Google BERT is the official pre-trained model from \cite{devlin2019bert}.

\vpara{RoBERTa-2seq \& RoBERTa-1seq} We compare with two versions of RoBERTa~\citep{liu2019roberta}. RoBERTa-2seq is trained with both masked language model~(MLM) task and next sentence prediction~(NSP) task, while RoBERTa-1seq refers to
the pre-training model with only the MLM task.

\vpara{\sparse{}}  We pre-train BERT models with its Transformer encoder replaced by a Sparse Transformer encoder~\citep{child2019generating}. We set its sparsity hyper-parameters stride $\ell=128$ and expressivity $c=32$.\footnote{We adopt Sparse Transformer implemented by Fairseq,
which first computes the $N\times N$ attention matrix, and then masks it to be a sparse one.
This implementation cannot avoid the $O(N^2)$ attention computation, and thus
has a similar training time/memory cost to RoBERTa.
}
The attention masking matrix used in Sparse Transformer and more implementation details are discussed in Appendix~\ref{subsec:sparse}. A similar architecture
was adopted in GPT-3~\cite{brown2020language}.

\subsection{Pre-training}

All the models follow the BERT-Base setting, i.e., $L=12, H=768, A=12$, and are trained on the same corpus --- BooksCorpus and English Wikipedia with uncased word piece tokens.
Thus all models use the same vocabulary as Google BERT~(uncased version) with vocabulary size 30,522.
We fix the number of tokens per batch $B\times N=131,072$, i.e., if sequence length $N=512$ then batch size $B=256$, if sequence length $N=1024$ then batch size $B=128$. The detailed pre-training configuration is listed in Appendix~\ref{subsec:config}.
Moreover, the pre-training of \sparse{} and \shortmethod{} follows the RoBERTa-1seq setting, i.e., we drop the NSP~(Next Sentence Prediction) task, and an input sequence is up to $N$ tokens until it reaches a document boundary.

\begin{table*}[tbp]
  \begin{center}
    \small
    \begin{tabular}{l |l| l| l|  l|  l}
      \toprule
      $N$                   & Model              & Training Time~(day) & Memory~(per GPU, GB) & Heads Config. & Valid. ppl \\\midrule
      \multirow{3}{*}{512}  & RoBERTa-1seq       & 6.62                & 9.73                 & -             & 3.58       \\
                            & \shortmethod{}~n=2 & 5.83~(-12.0\%)      & 7.91~(-18.7\%)       & 10:2          & 3.56       \\
                            & \shortmethod{}~n=3 & 5.80~(-12.5\%)      & 7.32~(-23.8\%)       & 8:2:2         & 3.71       \\\midrule
      \multirow{3}{*}{1024} & RoBERTa-1seq       & 9.66                & 13.39                & -             & 3.60       \\
                            & \shortmethod{}~n=2 & 7.51~(-22.3\%)      & 9.73~(-27.3\%)       & 9:3           & 3.57       \\
                            & \shortmethod{}~n=3 & 7.23~(-25.1\%)      & 8.55~(-36.1\%)       & 8:2:2         & 3.63       \\\bottomrule
    \end{tabular}
    \normalsize
      \caption{Pre-training Performance Analysis.}
  \label{tbl:pretrain}
  \end{center}
      \vspace{-0.1in}
\end{table*}
A summary of the pre-training performance comparison between \method{} and RoBERTa-1seq is shown in Table~\ref{tbl:pretrain}.
Besides memory saving, we also achieve a significant speedup. For example, when $N=1024$, \method{}~($n=2$) reduces the training time from RoBERTa's 9.7 days to 7.5 days.

\subsection{Fine-tuning Tasks}

We evaluate \method{} on several question answering tasks, including SQuAD~1.1/2.0~\citep{rajpurkar2018know} and five other tasks from the MrQA shared task\footnote{\url{mrqa.github.io}} --- HotpotQA~\citep{yang2018hotpotqa}, NewsQA~\citep{trischler2017newsqa}, SearchQA~\citep{dunn2017searchqa}, TriviaQA~\citep{joshi2017triviaqa} and NaturalQA~\citep{kwiatkowski2019natural}. Since MrQA does not have an official test set, we follow  \citet{joshi2019spanbert} to split the development set evenly to build a new development set and test set.

These QA datasets have different paragraph length distributions and are thus ideal for testing the effectiveness of \method{}\footnote{The detailed paragraph length distributions can be found in Appendix~\ref{subsec:para_length}}.
For example, SQuAD, NaturalQA, and HotpotQA consist of mostly short paragraphs~(shorter than 512), while paragraphs in SearchQA~(average length 1,004) and TriviaQA~(average length 934) have around 1,000 tokens.  
When the input sequence is longer than $N$, we 
follow the common practice~\cite{joshi2019spanbert} to split it using a sliding window of size $N$ and stride  128.
This means that for SearchQA and TriviaQA, a model with $N=512$ can only capture half of the context, while a model with $N=1024$ can accept the whole paragraph as input.

For all models, we adopt the same fine-tuning QA setup from \citet{devlin2019bert}. The tokenized paragraph $(p_1, \cdots, p_s)$ and question $(q_1, \cdots, q_t)$ are concatenated to be a sequence $\texttt{[CLS]} q_1 \cdots q_t \texttt{[SEP]} p_1 \cdots p_s \texttt{[SEP]}$. The sequence is then fed into the pre-trained model with two extra linear layers for predicting the start and end positions of the answer spans. The detailed fine-tuning setting is listed in Appendix~\ref{subsec:ft_config}.  Table~\ref{tbl:squad} and Table~\ref{tbl:mrqa} report the experimental results.

\begin{table}[htbp]
  \begin{center}
    \small
    \begin{tabular}{@{}l@{~}|@{~}l@{~}|  c  c | c  c@{}}
      \toprule
                           &                           & \multicolumn{2}{c|}{SQuAD 1.1} & \multicolumn{2}{c}{SQuAD 2.0}                                   \\
      $N$                  & Model                     & EM                            & F1                            & EM             & F1             \\\midrule
      -                    & Human Perf.               & 82.30                         & 91.20                         & 86.80          & 89.40          \\\midrule
      \multirow{7}{*}{512} & Google BERT               & 81.19                         & 88.45                         & 74.08          & 77.16          \\
                           & XLNet                     & -                             & -                             & 78.46          & 81.33          \\
                           & RoBERTa-2seq              & 82.91                         & 89.78                         & 75.79          & 79.17          \\
                           & RoBERTa-1seq              & \textbf{84.43}                & \textbf{91.48}                & \textbf{79.22} & \textbf{82.27} \\
                           & \sparse{}                 & 80.49                         & 88.09                         & 74.15          & 76.96          \\
                           & \shortmethod{}~n=2\hide{, 10:2}  & \textit{84.08}                & \textit{90.77}                & \textit{78.34} & \textit{81.46} \\
                           & \shortmethod{}~n=3\hide{, 8:2:2} & 82.37                         & 89.64                         & 77.33          & 80.33          \\ \midrule
      \multirow{4}{*}{1024}
                           & RoBERTa-1seq              & \textbf{84.58}                & \textbf{91.14}                & \textbf{79.34} & \textbf{82.26} \\
                           & \sparse{}                 & 81.02                         & 88.37                         & 74.51          & 77.57          \\
                           & \shortmethod{}~n=2\hide{, 9:3}   & \textit{83.65}                & \textit{90.74}                & \textit{78.55} & \textit{81.45} \\
                           & \shortmethod{}~n=3\hide{, 8:2:2} & 82.74                         & 90.05                         & 76.79          & 79.84          \\\bottomrule
    \end{tabular}
    \normalsize
      \caption{Dev set results on SQuAD~1.1/2.0. The result of XLNet(-Base) is from \citet{yang2019xlnet}. For \method{} models, their attention head configurations are the same as Table~\ref{tbl:pretrain}.}
  \label{tbl:squad}
  \end{center}
      \vspace{-0.1in}
\end{table}

\vpara{\method{}~(n=2) v.s. RoBERTa-1seq}
Comparing \method{} with RoBERTa-1seq when $N=512$, 
we observe an absolute F1 difference from
0.04~(in NaturalQA) to 1.18~(in NewsQA), with an average of 0.55. 
For $N=1024$,  \method{} achieves more comparable or even better performance to RoBERTa-1seq,
In SearchQA,  NewsQA and HotpotQA, \method{} achieves absolute F1 improvement of
0.39, 0.44 and 0.23, respectively.

\vpara{\method{} v.s. \sparse{}}
For $N=512$,  it is interesting  that \method{} with 3 blocks~(density 33.33\%)
performs better then \sparse{}~(density 44.20\%) in both SQuAD and MrQA tasks. 
Similar results can be observed for $N=1024$, too. These results show that off-diagonal masking matrices, e.g., the masking matrix defined by permutation $(2, 3, 1)$ and $(3,1,2)$,  play crucial roles in \method{}.
Furthermore, \method{} with 2 blocks achieve a more significant improvement.

\vpara{Effect of Long Sequence Pre-training} Our observations are twofold: (1) Long sequence pre-training benefits long sequence fine-tuning. In TriviaQA and SearchQA, of which paragraph lengths are around 1024, pre-training models with $N=1024$ achieve significantly better performance. (2) The heterogeneity of pre-training and fine-tuning sequence length may hurt performance. For example, in SQuAD, we do not see significant performance gain by using pre-trained models with $N=1024$; in HotpotQA and NewsQA, longer sequence pre-training even hurts performance.

\vpara{Effect of \#Blocks} It is not surprising  that \shortmethod{} with 2 blocks~($n=2$) performs better than that with 3 blocks~($n=3$), because it keeps more attention matrix entries.  The biggest difference is in SQuAD~2.0 and NewsQA with $N=1024$, where we observe an absolute loss of 1.6 F1 by increasing block number from 2 to 3.

\begin{table*}[htbp]
  \begin{center}
    \small
    \begin{tabular}{l|l|c c |c  c| c c| c c| c c}
      \toprule
                            &                           & \multicolumn{2}{c|}{SearchQA} & \multicolumn{2}{c|}{TriviaQA} & \multicolumn{2}{c|}{NewsQA} & \multicolumn{2}{c|}{NaturalQA} & \multicolumn{2}{c}{HotpotQA}                                                                                                                 \\
      $N$                   & Model                     & EM                               & F1                               & EM                             & F1                                & EM                               & F1                      & EM             & F1             & EM                      & F1                      \\
      \midrule
      \multirow{6}{*}{512}  & Google BERT               & 74.94                            & 80.37                            & 70.18                          & 75.35                             & 51.27                            & 66.25                   & 66.13          & 78.29          & 60.50                   & 77.08                   \\
                            & RoBERTa-2seq              & 76.12                            & 81.74                            & 71.92                          & 76.79                             & 52.45                            & 66.73                   & 66.98          & 78.63          & 61.52                   & 77.81                   \\
                            & RoBERTa-1seq              & \textbf{77.09}                   & \textbf{82.62}                   & \textbf{73.65}                 & \textbf{78.22}                    & \textbf{56.13}                   & \textbf{70.64}          & \textbf{67.14} & \textbf{79.07} & \textbf{62.77}          & \textbf{79.28}          \\
                            & \sparse{}                 & 73.36                            & 79.01                            & 68.71                          & 73.15                             & 51.18                            & 65.47                   & 65.53          & 77.46          & 58.54                   & 74.85                   \\
                            & \shortmethod{}~n=2\hide{, 10:2}  & \textit{76.68}                   & \textit{82.33}                   & \textit{72.36}                 & \textit{77.53}                    & \textit{54.66}                   & \textit{69.46}          & \textit{66.94} & \textit{79.03} & \textit{62.13}          & \textit{79.15}          \\
                            & \shortmethod{}~n=3\hide{, 8:2:2} & 75.54                            & 81.07                            & 72.05                          & 76.74                             & 53.82                            & 68.39                   & 66.14          & 78.47          & 60.64                   & 77.46                   \\
      \midrule
      \multirow{4}{*}{1024} & RoBERTa-1seq              & 77.47                            & 83.12                            & \textbf{75.29}                 & \textbf{80.20}                    & 55.00                            & 69.64                   & \textbf{68.28} & \textbf{80.35} & 61.89                   & 78.71                   \\
                            & \sparse{}                 & 74.83                            & 80.54                            & 70.56                          & 75.34                             & 51.67                            & 67.16                   & 65.07          & 77.31          & 59.65                   & 76.02                   \\
                            & \shortmethod{}~n=2\hide{, 9:3}   & \textit{\textbf{77.95}}          & \textit{\textbf{83.51}}          & \textit{75.06}                 & \textit{79.41}                    & \textit{\textbf{55.44}}          & \textit{\textbf{70.08}} & \textit{67.31} & \textit{79.39} & \textit{\textbf{62.13}} & \textit{\textbf{78.94}} \\
                            & \shortmethod{}~n=3\hide{, 8:2:2} & 76.98                            & 82.76                            & 74.78                          & 79.28                             & 53.48                            & 68.50                   & 65.91          & 78.20          & 61.89                   & 78.18                   \\
      \bottomrule
    \end{tabular}
    \normalsize
      \caption{MrQA test results~(Tasks are sorted decreasingly by average paragraph length). For \method{} models, their attention head configurations are the same as Table~\ref{tbl:pretrain}.}
  \label{tbl:mrqa}
  \end{center}
\end{table*}

\begin{table}[t]
  \centering
  \small
  \centering
  \begin{tabular}{@{}l@{~}|@{~}r @{~~~}r @{~~~}r @{~~~}r@{}}
    \toprule
    $B\times N$   & 8$\times$1024 & 16$\times$1024 & 24$\times$1024 & 32$\times$1024 \\ \midrule
    RoBERTa       & 0.1371        & OOM            & OOM            & OOM            \\
    \method{}~n=2 & 0.0990        & 0.1869         & OOM            & OOM            \\
    \method{}~n=3 & 0.0954        & 0.1790         & 0.2634         & OOM            \\
    \bottomrule
  \end{tabular}
  \normalsize
    \caption{Test time statistics~(sec) for different input size.
  OOM indicates out-of-memory.}
  \label{tbl:test}
\end{table}

\vpara{Efficient inference with \method{}}
We benchmark  test efficiency of RoBERTa and \method{}.
The benchmark code follows huggingface\footnote{\url{github.com/huggingface/transformers/blob/master/examples/benchmarks.py}}.  All experiments are run 30 times on a 32GB V100 GPU with half precision~(FP16). We report the average running time in Table~\ref{tbl:test}.
As we can see,  \method{} does achieve speedup and memory reduction during test time.
Take 8$\times$1024, i.e., batch size $B=8$, sequence length $N=1024$, as an example, we can see that \method{} with 2 blocks saves 27.8\% of test time, and \method{} with 3 blocks saves more~(30.4\%). As for memory, we can observe that RoBERTa cannot handle an input of size 16$\times$1024, while it is possible for \method{} to work on it.

In summary, not only \method{} saves training/inference time and  memory, but it also has a competitive and sometimes better performance, especially for tasks with longer sequences.
This demonstrates the effectiveness of our blockwise multi-head attention approach.

\subsection{Ablation Study}
\label{sec:ablation}

\begin{figure*}[htbp]
     \centering
     \subfigure[$N=512, n=2$]{
        \includegraphics[width=.23\textwidth]{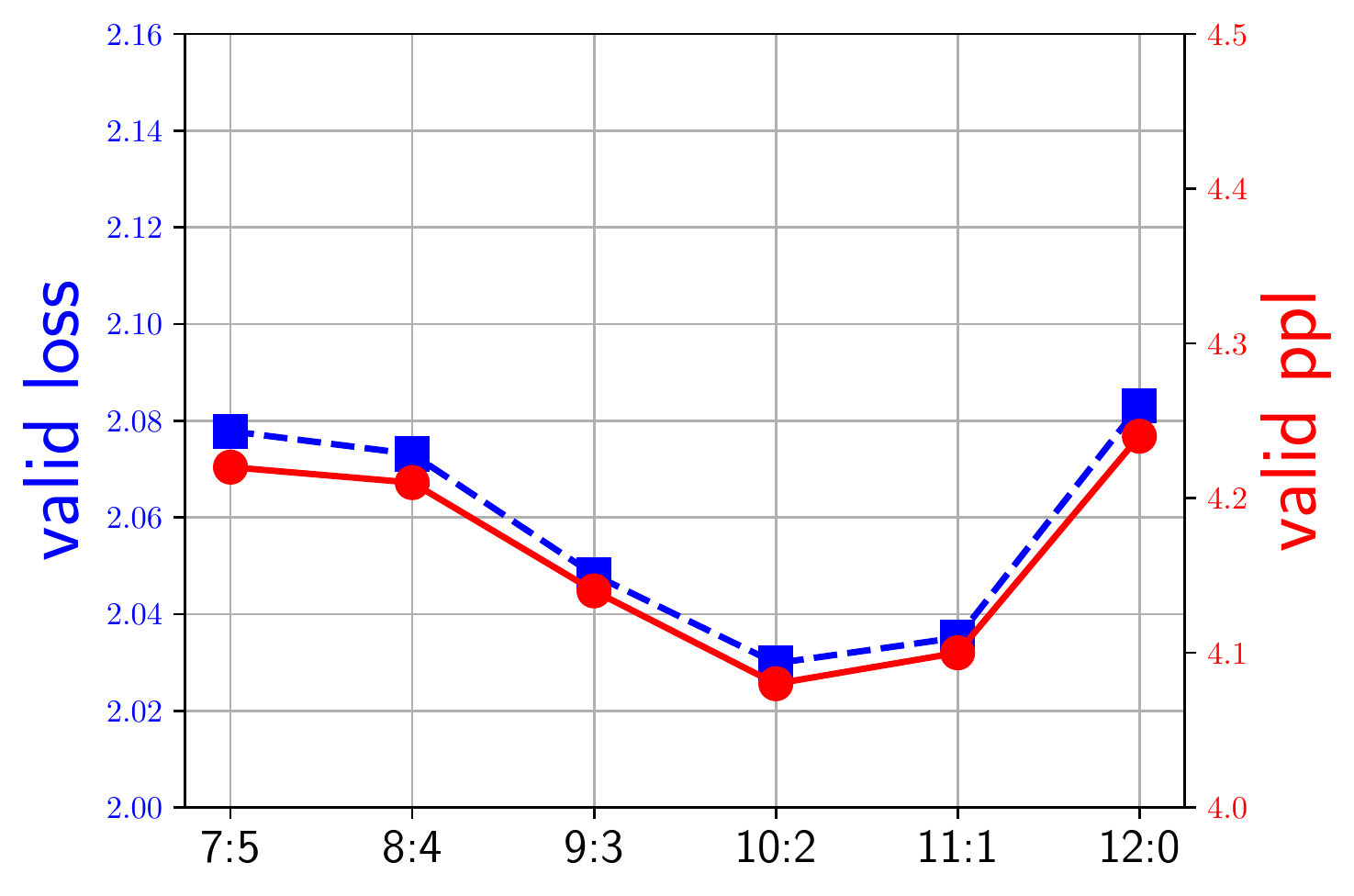}
        \label{fig:ablation_len_512_block_2}
        }
     \subfigure[$N=1024, n=2$]{
         \includegraphics[width=.23\textwidth]{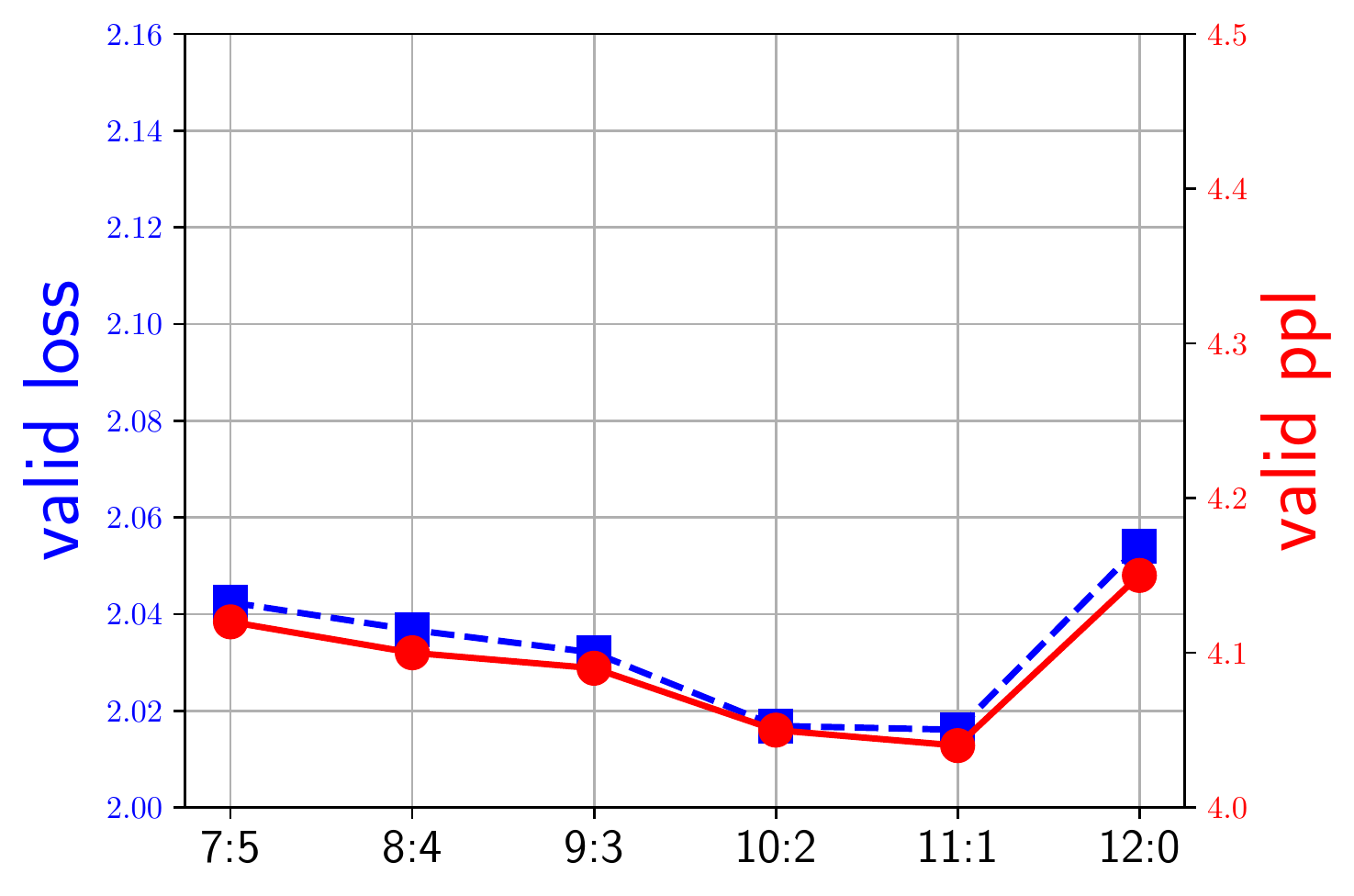}
        \label{fig:ablation_len_1024_block_2}
        }
     \subfigure[$N=512, n=3$]{
        \includegraphics[width=.23\textwidth]{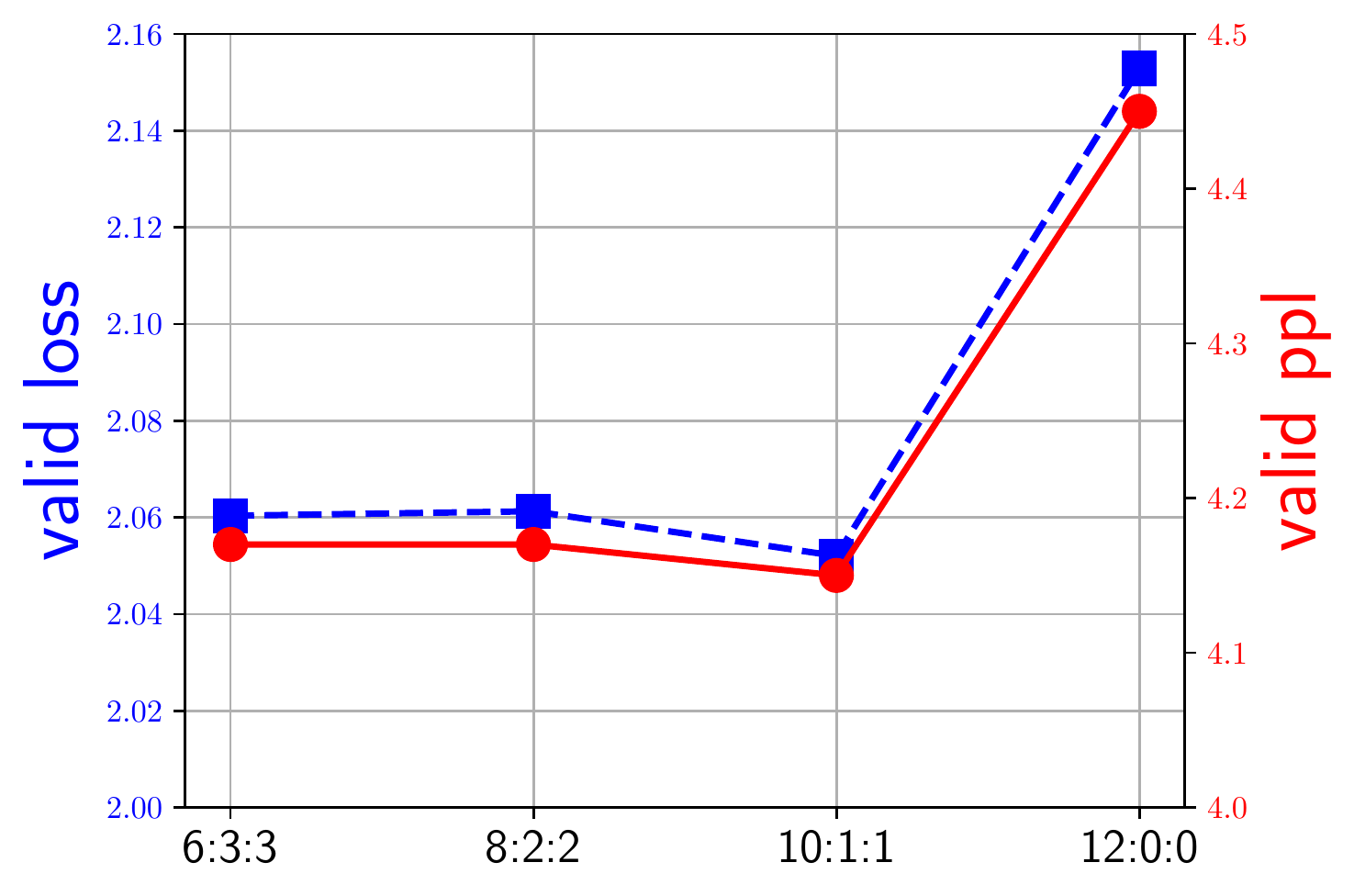}
        \label{fig:ablation_len_512_block_3}
        }
     \subfigure[$N=1024, n=3$]{
         \includegraphics[width=.23\textwidth]{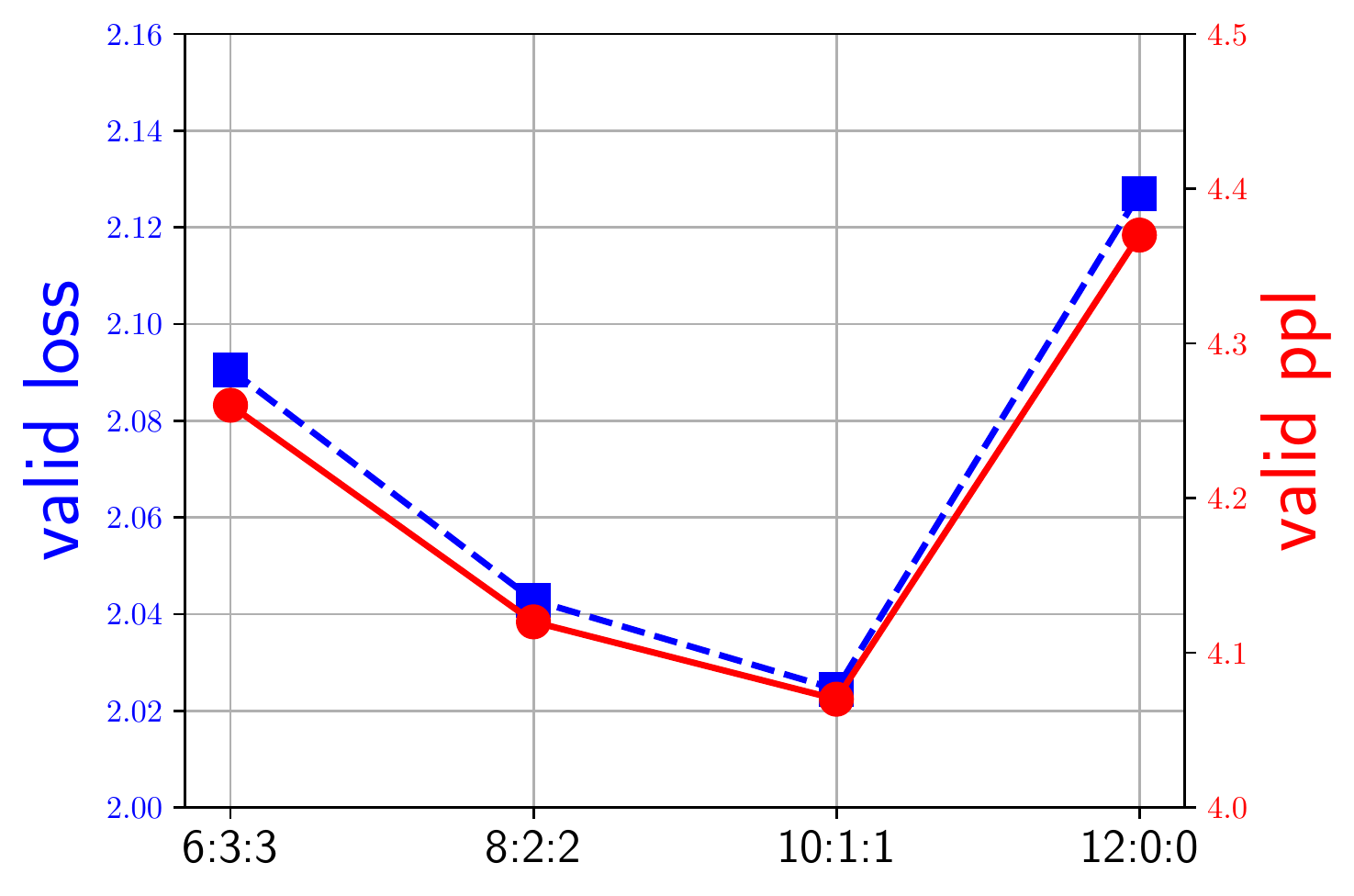}
        \label{fig:ablation_len_1024_block_3}
        }
     \caption{Ablation over blockwise attention heads assignment. 
     }
     \label{fig:ablation}
\end{figure*}

We fix the assignment of attention heads in  the above experiments. For example, \method{} with sequence length $N=512$ and 2 blocks is trained with ten heads using permutation $(1, 2)$ and the other two using permutation $(2, 1)$. However, there are other ways to assign twelve attention heads, e.g., seven heads for permutation $(1, 2)$ and the other five for permutation $(2, 1)$.
It would be interesting to see how the assignment of heads affects model performance. In this section, we grid search attention head assignments and plot their best validation performance in 1.2M training steps. The results are shown in \Figref{fig:ablation}.

Our observations are threefold: (1) Identity permutations, i.e., $(1, 2)$ and $(1, 2, 3)$, are important. As shown in \Figref{fig:ablation}, all optimal solutions assign considerable attention heads to block-diagonal matrices, since those matrices enable  each token to attend to its nearby tokens;
(2) Non-identity permutations follow the rule of  ``vital few and trivial many."  Although identity permutations are important, assigning all attention heads to them~(corresponding to 12:0 and 12:0:0 in \Figref{fig:ablation}) significantly hurts performance, since the model can not learn long-term dependencies with only identity permutation;
(3) Pre-training performance and fine-tuning performance are correlated but not always consistent. 
When $n=3$, pre-training performance suggests 10:1:1 to be the best head assignment --- ten heads for permutation $(1,2,3)$, one head for  $(2, 3, 1)$ and one head for $(3, 1, 2)$, but we observe that the configuration of 8:2:2 achieves better performance in fine-tuning tasks.

\section{Related Work}
\label{sec:related}

In this section, we review the related work of memory optimization for neural network training and recent efforts to simplify Transformer and BERT. 

\subsection{Low-memory neural networks training}

Due to the large size of model parameters and deep architectures, modern neural networks training requires significant amounts of computing resources.
As a result, there is an increasing interest in training neural networks with low memory~\citep{sohoni2019low}. 
Mainstream techniques mostly address this problem with a better system or engineering design, such as  
low-precision training~\citep{micikevicius2017mixed},  microbatching~\citep{huang2018gpipe} and gradient checkpointing~\citep{chen2016training}.
Alternatively, there also exists some research focusing on the theoretical aspect, including the recently proposed lottery ticket hypothesis~\citep{frankle2018lottery}.

\subsection{Efficient Transformer}
\label{subsec:discussion}
Since the invention of Transformer~\citep{vaswani2017attention} and its successful application to masked language model 
pre-training~\citep{devlin2019bert, radford2019language, yang2019xlnet, liu2019roberta,lan2019albert},
several approaches have been proposed to simplify the model and its training process.  
We summarize these attempts as follows:

\vpara{Attention layer simplification}
There are currently two lines of research trying to simplify the multi-head attention layers.
The first one focuses on attention matrix sparsification.
Notable examples include Star Transformer~\citep{guo2019star}, Sparse Transformer~\citep{child2019generating}, Adaptive Sparse Transformer~\citep{correia2019adaptively, sukhbaatar2019adaptive},  Log-Sparse Transformer~\cite{li2019enhancing} , Reformer~\citep{kitaev2020reformer} and Longformer~\citep{beltagy2020longformer}.
However, due to the insufficient support for sparse tensors from the current deep learning platforms, some of them have to represent a sparse matrix using a dense matrix with a binary mask or rely on customized CUDA kernels~\citep{gray2017gpu}.
As a result, the speed-up or reduction in memory consumption is sometimes limited in practice.
The second line of research prunes redundant attention heads.
Examples include \cite{voita2019analyzing} and \cite{michel2019sixteen}.
Our \method{} model belongs to the first category, as we sparsify the attention matrices to be 
block sparse matrix.

\vpara{Reducing model size for pre-training}
Knowledge distillation~\cite{hinton2015distilling} is a general technique that aims to compress and transfer knowledge from a teacher model to a simpler student model.
There are two recent efforts that apply knowledge distillation to BERT pre-training for reducing model size:
TinyBERT~\cite{jiao2019tinybert} distills BERT using a smaller Transformer, and \citet{tang2019distilling} distills BERT with a BiLSTM.
In contrast, ALBERT~\cite{lan2019albert} is a notable work that does not take the knowledge distillation approach.
It uses parameter-sharing to reduce the number of parameters of the BERT model.
As discussed in \secref{subsec:profile}, parameter-sharing reduces both model memory and optimizer memory.
These two parts account for about 12.4\% of total training memory for BERT-base.
As for efficiency, parameter-sharing reduces communication complexity in distributed training and thus saves training time as well.
\hide{
\begin{table}[t]
    \caption{Comparison between Sparse Transformer and Transformer-XL on Enwik8. Results are from~\citet{child2019generating}.}
    \label{tbl:enwik8}
    \small
    \centering
    \begin{tabular}{l|r}
      \toprule
      Model                                         & Bits per byte \\\midrule
      Transformer-XL~88M~\cite{dai2019transformer}  & 1.03          \\
      Transformer-XL~227M~\cite{dai2019transformer} & 0.99          \\
      Sparse Transformer~95M~(fixed)                & 0.99          \\
      \bottomrule
    \end{tabular}
    \normalsize
  \end{table}
  }

In the aforementioned efficient Transformers, the model quality is often demonstrated by comparable language model perplexity, or equivalently the bits per word/byte. 
It is often implicitly assumed that similar language model perplexity implies similar pre-training model quality, namely the same performance on the downstream tasks.
We would like to point out that this assumption does not necessarily hold.  
For example, 
the experiments on the Enwik8 dataset by \citet{child2019generating} demonstrates that Sparse Transformer  ``surpasses the 1.03 state-of-the-art (bits per byte) for
a similarly-sized Transformer-XL and matching the 0.99 (bits per byte) of a model trained with more than double the number of parameters''.
However, if we compare \sparse{}~(pre-training model with Sparse Transformer backbone) against XLNet~\cite{yang2019xlnet}~(pre-training model with Transformer-XL backbone) in SQuAD, Table~\ref{tbl:squad} shows that XLNet still outperforms 
\sparse{} significantly.
Therefore, we believe that it is necessary to conduct a comprehensive study and evaluation of existing efficient Transformer models when used for masked language model pre-training.
Limited by resources, in this work, we mainly compare \method{} to pre-training using Sparse Transformer~\cite{child2019generating}, which is the earliest attempt to design efficient Transformer models
and also the key contributor to the success of GPT-3~\cite{brown2020language}.
We plan to benchmark more models in the future.
\section{Conclusion}
\label{sec:conclusion}

In this work, we study the lightweight BERT model with the goal of achieving 
both efficiency and effectiveness. We profile and analyze the memory bottlenecks of BERT and focus on optimize dot-product self-attention, which consumes quadratic memory with respect to the sequence length. 
To reduce both time and memory consumption, we present \method{}, 
which sparsifies the attention matrices to be sparse block matrices. 
The proposed model achieves time and memory saving without significant loss of performance. 

In the future,  we plan to
benchmark more efficient Transfomers in language model pre-training and fine-tuning.
We also would like to explore more applications of \method{} on NLP tasks involving long sequences such as coreference resolution~\citep{joshi2019bert} and document-level machine translation~\citep{miculicich2018document}, 
and also non-NLP tasks such as protein sequence modeling~\citep{rives2019biological, rao2019evaluating}.

\section*{Acknowledgments}
The authors would like to thank 
Zhilin Yang, Danqi Chen, Yinhan Liu, Mandar Joshi and  Luke Zettlemoyer for the helpful suggestions.
Jiezhong Qiu and Jie Tang were partially supported by the
National Key R\&D Program of China (2018YFB1402600),
NSFC for Distinguished Young Scholar 
(61825602),
and NSFC (61836013).


\newpage
\bibliographystyle{acl_natbib}
\bibliography{reference}

\appendix
\section{Appendix}

\subsection{Notations and Pre-training Hyper-parameters}
\label{subsec:config}

The notations and pre-training hyper-parameters are listed in Table~\ref{tbl:notation} and Table~\ref{tbl:hyper}.

\begin{table}[h]
  \small
  \centering
  \begin{tabular}{l  l  l  l}
    \toprule
         & Description                  & Base & Large \\\midrule
    $B$  & Batch size                   & 256  & 256   \\
    $A$  & \# Self-attention heads      & 12   & 16    \\
    $L$  & \# Layers                    & 12   & 24    \\
    $H$  & \# Hidden units              & 768  & 1024  \\
    $4H$ & \# Feed-forward hidden units & 3072 & 4096  \\
    $N$  & Sequence length              & 512  & 512   \\\bottomrule
  \end{tabular}
  \normalsize
    \caption{BERT notations.}
  \label{tbl:notation}
\end{table}

\begin{table}[h]
  \centering
  \small
  \begin{tabular}{l l}
    \toprule
    Hyper-parameter       & Value   \\\midrule
    Vocabulary Size       & 30,522  \\
    Dropout               & 0.1     \\
    Attention dropout     & 0.1     \\
    Warmup steps          & 10K     \\
    Weight decay          & 0.01    \\
    Max steps             & 2.4M    \\
    Initial learning rate & 0.00025 \\
    Learning rate decay   & Linear  \\
    Adam $\epsilon$       & 1e-8    \\
    Adam $\beta_1$        & 0.9     \\
    Adam $\beta_2$        & 0.999   \\
    Gradient Clipping     & 1.0     \\\bottomrule
  \end{tabular}
  \normalsize
    \caption{Pre-training hyper-parameters.}
  \label{tbl:hyper}
\end{table}

\subsection{Profiler Implementation}
\label{subsec:profiler}

Among the three types of training memory, model memory and optimizer memory is relatively easy to profile~(can be computed by enumerating each tenor and summing up \texttt{tensor.numel() * tensor.element\_size()}).
To calculate activation memory, \cite{sohoni2019low} traverse PyTorch's autograd graph and sum up the necessary storage space. They find that the summation of model memory, optimizer memory, and activation memory matches PyTorch memory profiling tool\footnote{
\texttt{torch.cuda.max\_memory\_allocated}}. 

Based on their observation, we use the following quantity as an estimate to activation memory
\beq{
  \label{eq:activation}
  \text{max memory allocated}-\text{model memory}-\text{optimizer memory}
} When profiling BERT, we first pre-train it for 1000 steps, and then compute its model and optimizer memory. Finally, we estimate its activation memory according to \Eqref{eq:activation}.

\subsection{\sparse{}}
\label{subsec:sparse}

\begin{figure}[htbp]
  \centering
  \subfigure[]{
    \includegraphics[width=.22\textwidth]{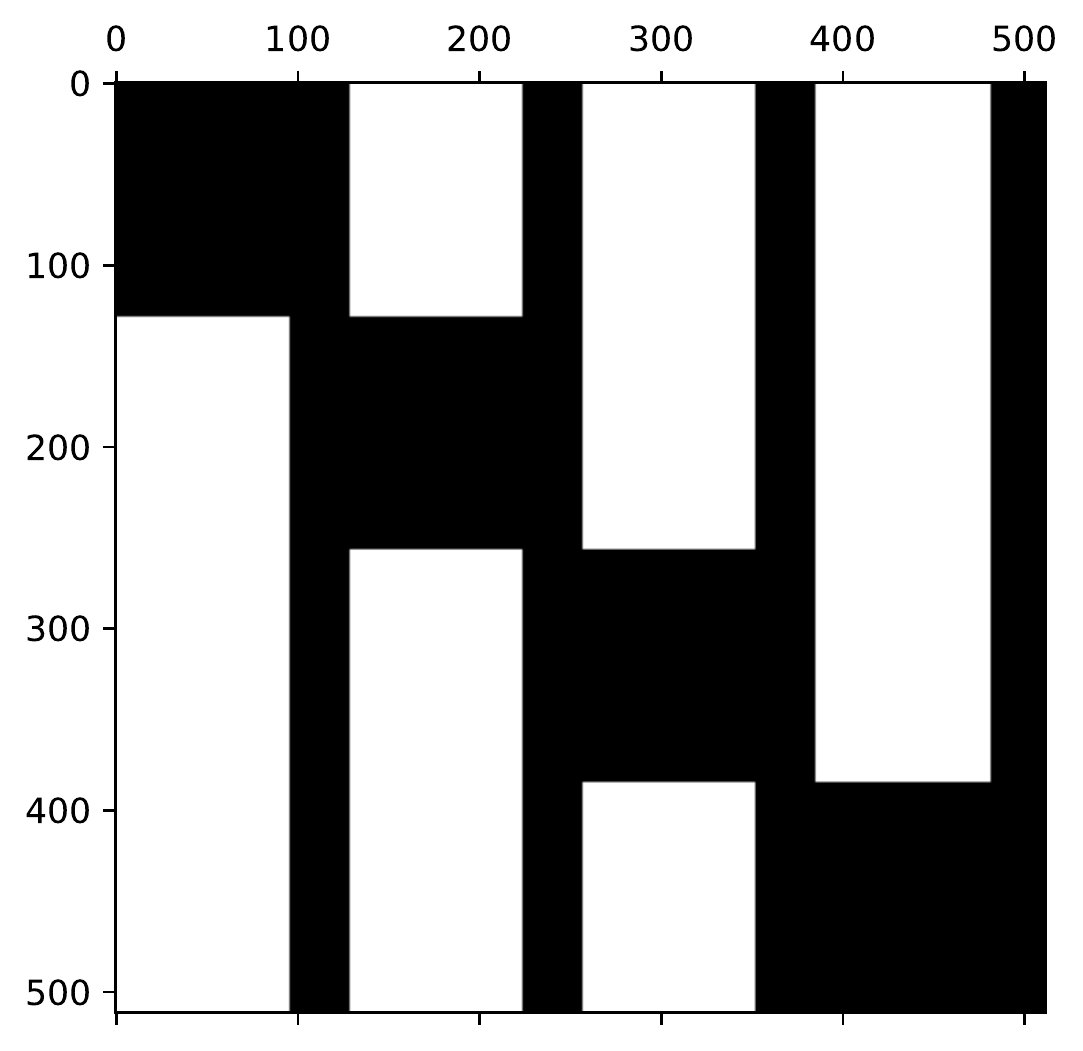}
    \label{fig:sparse_512_128_32}
  }
  \subfigure[]{
    \includegraphics[width=.22\textwidth]{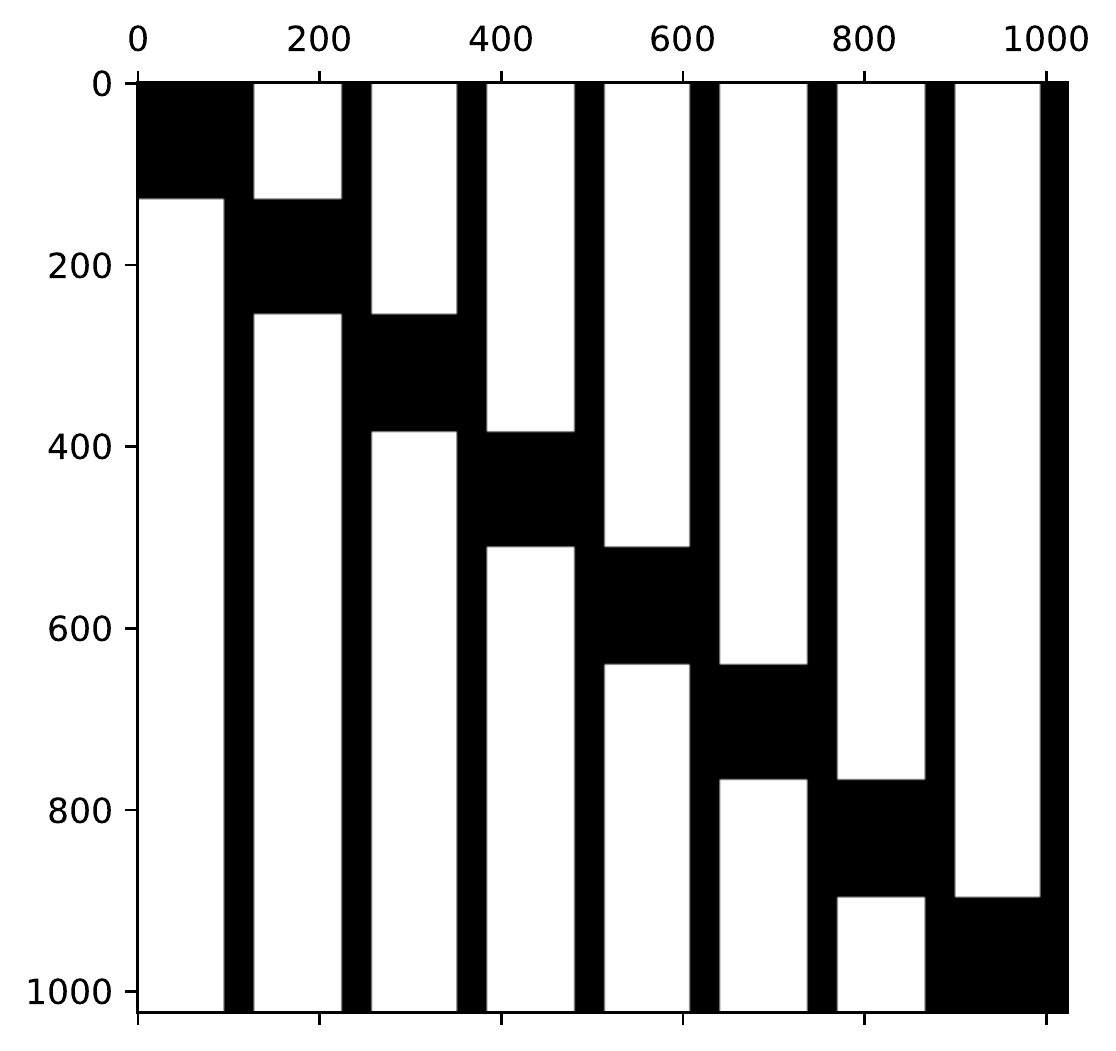}
    \label{fig:sparse_1024_128_32}
  }
  \caption{The sparse masking matrices we use in Sparse Transformer~(fixed mode) encoder. White color indicates attention values to be masked.
    (a) $N=512, \ell=128, c=32$, density 44.20\%;
    (b) $N=1024, \ell=128, c=32$, density 34.97\%. }
  \label{fig:sparse_mask}
\end{figure}

The sparse masking matrices we use for Sparse Transformer~\citep{child2019generating} are shown in \Figref{fig:sparse_mask}. We adopt the implementation of Sparse Transformer from Fairseq\footnote{\url{github.com/pytorch/fairseq/blob/master/fairseq/modules/sparse\_multihead\_attention.py.}}. The Fariseq version is implemented in a direct way, with the goal of comparing performance, not speed. We first compute the $N^2$ attention matrix and then mask it to be a sparse matrix according to the sparse pattern defined in Sparse Transformer paper. Consequently, this implementation of SparseBERT has very close training time/memory cost as RoBERTa (as it can not avoid the $O(N^2)$ attention computation). We did so because the code released by Sparse Transformer is based on Tensorflow and relies on customized CUDA kernels, but our pre-training is done using PyTorch. 

\subsection{Fine-tuning Settings}
\label{subsec:ft_config}

Our fine-tuning is implemented based on code base from HuggingFace\footnote{\url{github.com/huggingface/pytorch-transformers}} and SpanBERT~\citep{joshi2019spanbert}.
We use \texttt{max\_sequence\_length=$N$}, i.e., we allow fine-tuning task to input sequences as long as the pre-training model. If the input sequence is too long to fit the \texttt{max\_sequence\_length=$N$} constraints, we use a sliding window of stride 128 to split it. We grid search learning rate from \{5e-6, 1e-5, 2e-5, 3e-5, 5e-5\} and batch size from \{16, 32\}. The fine-tuning is performed for 4 epoches.

\subsection{Paragraph-Length Distribution}
\label{subsec:para_length}

\begin{figure}[h]
  \begin{center}
    \includegraphics[width=.4\textwidth]{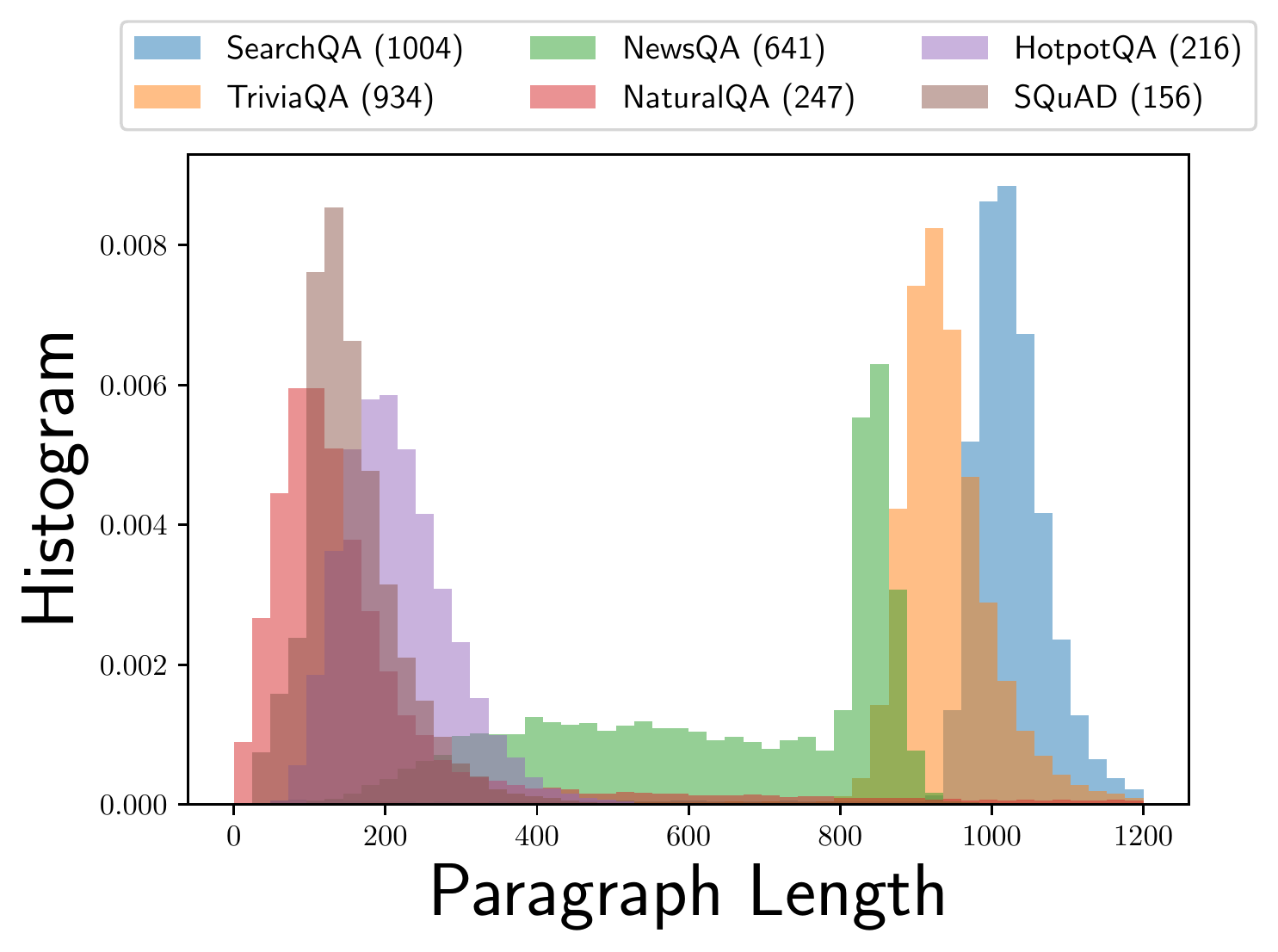}
  \end{center}
  \caption{Paragraph-length~(after tokenization) distribution. The distribution of SQuAD 2.0 is very similar to SQuAD 1.1, so we only plot SQuAD 1.1 here.}
  \label{fig:doc_length}
\end{figure}

The paragraph-length distribution
of SQuAD and MrQA datasets
is shown in \Figref{fig:doc_length}.

\end{document}


\twocolumn[
\icmltitle{
Supplementary Material of Blockwise Self-Attention for Long Document Understanding
}



\icmlsetsymbol{equal}{*}

\begin{icmlauthorlist}
\end{icmlauthorlist}

\icmlcorrespondingauthor{}{}

\icmlkeywords{Machine Learning, ICML}

\vskip 0.3in
]



\printAffiliationsAndNotice{}  

\appendix
\input{appendix.tex}

\bibliography{reference}
\bibliographystyle{icml2020}